\documentclass[runningheads]{llncs}
\usepackage{url}
\usepackage{algorithm}
\usepackage{algpseudocode}
\usepackage{graphicx}
\usepackage{multirow}
\usepackage{amsmath}
\usepackage{tabularx}
\usepackage{caption}
\usepackage{subcaption}
\usepackage{amssymb}

\begin{document} 
\title{Predicting Individual Substance Abuse Vulnerability using Machine Learning Techniques}

\titlerunning{Predicting Individual Substance Abuse Vulnerability}

\author{Uwaise Ibna Islam$^{1,*}$ \and Iqbal H. Sarker$^{1,*}$ \and Enamul Haque$^2$ \and Mohammed Moshiul Hoque$^1$
}
\authorrunning{Islam et al.}
\institute{$^1$ Dept of Computer Science \& Engineering, Chittagong University of Engineering \& Technology, Chittagong-4349, Bangladesh.\\
$^2$ McMaster University, Canada.\\
$*$Correspondence: iqbal@cuet.ac.bd
}
\maketitle            
\begin{abstract}
Substance abuse is the unrestrained and detrimental use of psychoactive chemical substances, unauthorized drugs, and alcohol. Continuous use of these substances can ultimately lead a human to disastrous consequences. As patients display a high rate of relapse, prevention at an early stage can be an effective restraint. We therefore propose a binary classifier to identify any individual’s present vulnerability towards substance abuse by analyzing subjects' socio-economic environment. We have collected data by a questionnaire which is created after carefully assessing the commonly involved factors behind substance abuse. Pearson’s chi-squared test of independence is used to identify key feature variables influencing substance abuse. Later we build the predictive classifiers using machine learning classification algorithms on those variables. Logistic regression classifier trained with 18 features can predict individual vulnerability with the best accuracy. 
\end{abstract}

\section{Introduction}
\label{Introduction}
Substance abuse refers to the harmful or hazardous use of psychoactive substances, including alcohol and illicit drugs \cite{worldhealthorganization}. It is widely regarded as one of the most alarming problems in the world \cite{degenhardt2012extent}. About 4.9\% adult population of the whole world, roughly 240 million people have been identified with alcohol abuse disorder. Among them 7.8\% of men and 1.5\% of women, with alcohol causing an estimated 257 disability-adjusted life years lost per 1,00,000 population. Around 22.5\% of adults all over the world, nearly a billion people smoke tobacco-related products which is a significant percentage in the global male population 32.0\% and a lesser yet large percentage of women 7.0\% \cite{gowing2015global}. Substance abuse varies from country to country, based on demographic factors, demands, and other issues. Prolonged abuse of these substances can be fatal, from 1999 to 2017 more than 3,99,000 people died from an overdose in any form of opioid \cite{scholl2019drug}. Alarmingly, opioid is just one of the most abused substances all around the world, other popularly abused substances are alcohol, amphetamine, cannabis, and nicotine. 

\par Substance abuse disorder is caused by the changes in the reward system of the brain from continuous abuse of substances \cite{kelley2002neuroscience}. As substance abuse disorder shows a tendency of relapse \cite{brandon2007relapse}, it is ideal to focus on prevention for solving the problem. So, in this study, we propose the prevention model focusing on the primary risk factors contributing to substance abuse. According to studies, this is the promising route to effective strategies of building a prevention model \cite{hawkins1992risk}. Reviewing relevant literature and discussing with experienced local psychiatrists, we find the following risk factors contributing the most to substance abuse disorder: Peer influence  \cite{oetting1987peer}\cite{sani2010drug}, Depression \cite{conner2009meta}, Stress \cite{sinha2008chronic}, Family relations \cite{barrett2006family}\cite{bahr1998family}, Occupational failure and/or unemployment \cite{nagelhout2017economic}, Curiosity \cite{sani2010drug}\cite{pierce2005role}, and Religious affiliations \cite{bahr1998family}.

Typically, to build a machine learning model or data-driven real world systems, the availability of data is the key  \cite{sarker2020cybersecurity} \cite{sarker2020mobile}. In our approach, we have collected data from individuals through a questionnaire, prepared from aforementioned factors. After pre-processing the data-set and calculating the significance of the feature variables on the target variable using Pearson's chi-squared test of independence, we have identified key features to train and build the classifier using machine learning algorithms. We take into account the most popular machine learning classification algorithms such as Logistic Regression, Decision Tree, Random Forest, K-nearest Neighbors, Support Vector Machines, and Naive Bayes classifier \cite{sarker2019effectiveness}. To select the optimal binary classifier we evaluate and compare performance between the classifiers. The contributions of this study can be summarized as follows:
\begin{itemize}
    \item Identifying and validating the key risk factors behind substance abuse.
    \item Predicting the vulnerability of any individual towards substance abuse using machine learning classification algorithms to facilitate prevention.
\end{itemize}

In the rest of the paper, we discuss related works in Section \ref{Related works}, present the methodology in Section \ref{Methodology}, evaluation and efficacy of the classifier are discussed in Section \ref{Evaluation and experimental results}, and conclude in Section \ref{Conclusion}.

\section{Related works}
\label{Related works}

Lots of studies have been done in the field of substance abuse disorder, studies related to ours are mentioned in this section. M. N. Sani et al. has analyzed the factors responsible, types of drugs being consumed, percentage of students receiving treatment and displaying relapsing behaviour in drug addiction among undergraduate students of private universities from 160 samples \cite{sani2010drug}. M. Aldarwish et al. has tried to find depression among the youth population by analyzing social media with a goal to classifying individuals examined as either depressed or not depressed using support vector machine and naive bayes classifiers \cite{aldarwish2017predicting}. S. Hassanpour et al. has build a model to identify any individual’s risk towards the use of alcohol, tobacco, and drugs using deep CNN (convolutional neural network) on images and LSTM (long short-term memory) on the text collected from the Instagram profiles of users \cite{hassanpour2019identifying}. This model however, is more successful in identifying alcohol compared to other substances. L. Fisher et al. has predicted certain personality traits which can be responsible for relapse after receiving treatment for substance abuse examining 108 patients using a 5-factor model of personality used in the neo-personality inventory \cite{fisher1998predicting}. 

\par In \cite{sani2010drug}, various aspects of drug addiction are analysed among the private university students who lives off-campus unlike residential students of public universities who are more prone to substance abuse as A.B. Heydarabadi et al. suggests \cite{heydarabadi2015prevalence}, so this study shows a partial representation of drug addiction among the undergraduates in Bangladesh. In our proposed study, we have collected data from respondents with diverse backgrounds. Moreover, unlike this study, apart from analysing, our proposed study searches for a solution to address the issue as well. In \cite{aldarwish2017predicting}, precision and recall values are not satisfactory as people often post in social media with double meaning and all users are not equally expressive about personal life on social media, making it difficult for this classifier to identify actually depressed users. In \cite{hassanpour2019identifying} also, Instagram data is used to identify individual risk, having similar problems of study \cite{aldarwish2017predicting}. In our proposed study, we have collected actual data of different patients diagnosed with substance abuse disorder from rehabilitation centers, providing diverse and accurate data to our classifier solving the problems regarding both these studies. In \cite{fisher1998predicting}, authors work on preventing relapse and offers very specific personal traits responsible for the relapse whereas we focus on prevention at a very early stage which is always a better option. In addition, we intend to incorporate people from random and diverse background to build an inclusive predictive classifier. 

\section{Methodology}
\label{Methodology}
\subsection{Preparing Questionnaire}
Discerning the risk factors responsible for the problem is the initial task. We have reviewed relevant literature to identify common risk factors. As these factors vary globally, we have corroborated our findings from relevant literature by local psychiatrists and added factors exclusive to Bangladesh, e.g., religious mindset is identified as a key factor in Bangladesh compared to other countries. The below 36 multiple choice questions based on these factors forms the questionnaire: \\
Q1. What is your age? \\
Q2. What is your Gender? \\
Q3. What is your Marital status? \\
Q4. What is your Relationship status right now? (if unmarried) \\
Q5. Do you belong to a broken family? (parents live separately) \\
Q6. Are you happy about your marriage or relationship? \\
Q7. what is the financial situation of your family? \\
Q8. Is or was any of your family members ever involved in drugs? \\
Q9. How is your relationship with your parents? \\
Q10. Do your family often curse you for failing to fulfil their expectations? \\
Q11. How successful are you among your family members? \\
Q12. Who are you living with right now? \\
Q13. What is your present occupational status? \\
Q14. How do you rate your occupational success? \\
Q15. How do you rate your workplace? \\
Q16. What's your occupational status compared to friends? \\
Q17. What's the number of your friends or accomplices? \\
Q18. Do you get influenced by your friends' activity? \\
Q19. How many of your friends have taken drugs so far? \\
Q20. Do you like to hang out with friends? \\
Q21. Do you join parties or hang-outs where substance is abused frequently? \\
Q22. How fast can you get into a society or community? \\
Q23. Do you have curiosity or fantasy about drugs? \\
Q24. Do you have access to drugs(could you manage if you wanted)? \\
Q25. Do you hate a drug addict as a person? \\
Q26. Are you satisfied with your life until now? \\
Q27. Are you happy with your physical outlook? \\
Q28. Are you suffering from any serious illness? \\
Q29. Have you lost any of your closed ones recently? \\
Q30. How do you rate your living place? \\
Q31. What is your religious mindset? \\
Q32. Are you involved in sports or any form of physical exercise? \\
Q33. Have you ever failed in love? (break-up, refusal of love) \\
Q34. Do you enjoy taking risks or having new experiences? \\
Q35. Do you think that people don't find interest in conversation with you?\\ 
Q36. Do you believe in the notion that drugs can relieve you from mental stress?
\par
\begin{figure}[h]
     \centering
    \includegraphics[width=0.9\linewidth]{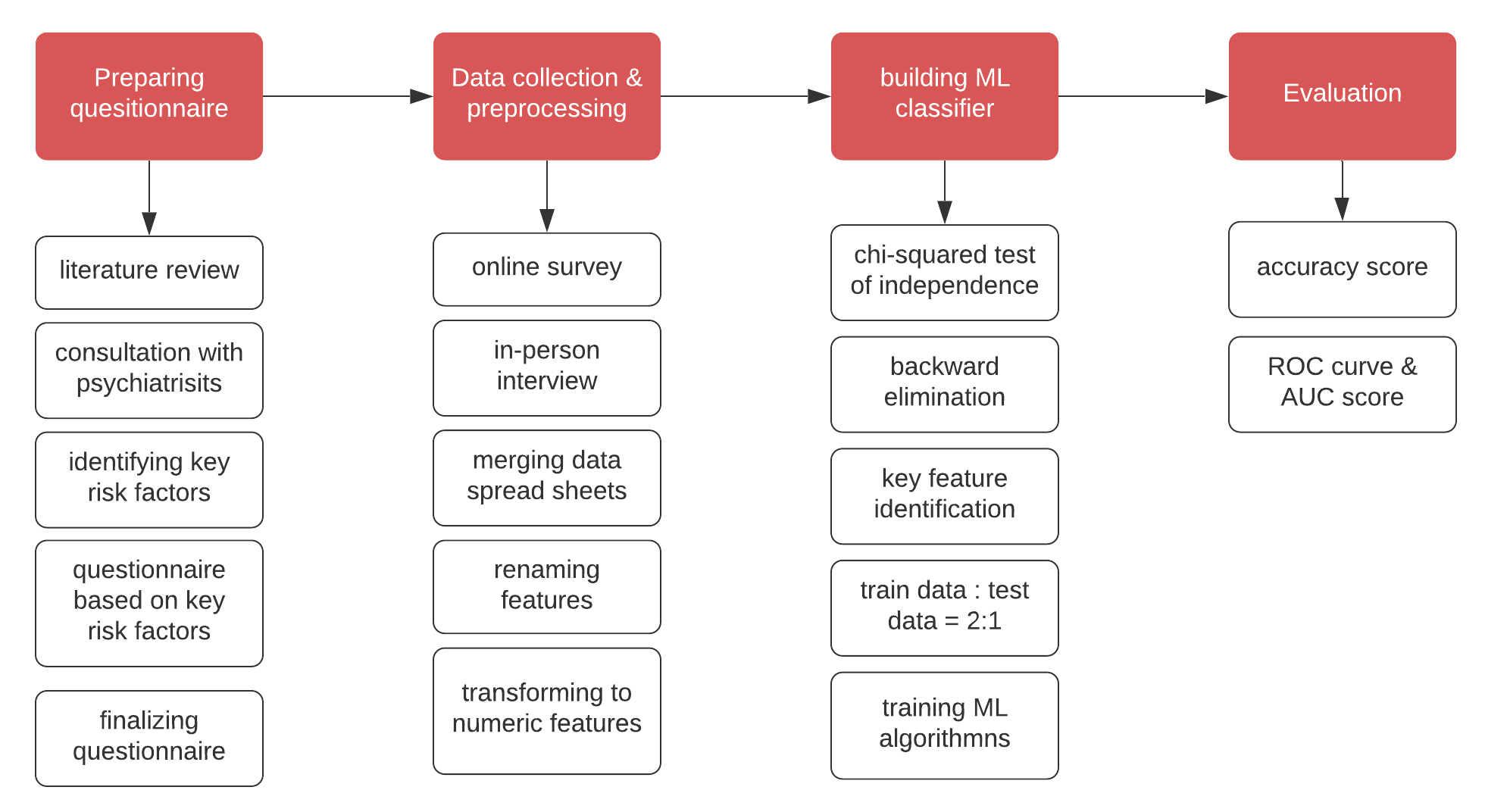}
    \caption{Methodology and Evaluation}
   \label{fig:methodology_&_evaluation}
\end{figure}

These questions are based on the key factors : peer influence, family and relationship, curiosity, personality traits, career success and unemployment and religious mindset alongside some introductory questions. We have examined the ability of the questionnaire to extract required information by pretesting among 25 respondents and later finalized it for data collection as in Figure \ref{fig:methodology_&_evaluation}.
\par 
\par
\subsection{Data collection and pre-processing}
We have collected individual data from both patients of substance abuse and healthy people for the same questionnaire. Answers to the identical questions contrasting the most between the two groups are considered as the key factors behind substance abuse in our study. Responses from healthy people have been collected through online survey. As for the patients, we have collected data by in-person interview from rehabilitation centres across Dhaka city due to the scarcity of voluntary responses concocted by social stigma attached to the problem. Answers to the questions are categorical: largely ordinal, some discrete and binary. 8 questions out of 36 are binary questions with two answers to choose from (Questions: 2, 4, 5, 21, 25, 32, 34, 35), remaining ones are 3-point Likert scale questions (28). These answers (n=486) from both groups are merged into a single data-set. Answers to these questions are categorical which have we later transformed into numerical values that helps training the algorithms afterwards \cite{sarker2020context}.

\subsection{Building Predictive Model}
We use Pearson's chi-squared test of independence to calculate the dependency between each of the feature variable and the target variable \cite{mchugh2013chi}. Before chi-squared statistic, $\chi^{2}$ is calculated, it has a null hypothesis that all the variables are independent of one another. The chi-squared test calculates the dependency by creating a contingency table of the two variables. If the table has $R$ number of rows and $C$ number of columns the chi-squared test of independence statistic, $\chi^{2}$ is given by the equation \cite{mchugh2013chi}:

\begin{equation*}
\chi^{2} = \sum_{i=1}^{R}{\sum_{j=1}^{C}{\frac{(o_{ij} - e_{ij})^{2}}{e_{ij}}}}   
\end{equation*}
\par 
here, $o_{ij}$ is the observed cell count in the ith row and jth column of the table and $e_{ij}$ is the expected cell count in the $i$th row and $j$th column of the table. Expected cell count $e_{ij}$ is calculated by the equation \cite{mchugh2013chi}: 
\begin{equation*}
    e_{ij} = \frac{\mathrm{ \textrm{row } \mathit{i}} \textrm{ total} * \mathrm{\textrm{col } \mathit{j}} \textrm{ total}}{\textrm{grand total}}  
\end{equation*}
The calculated $\chi^{2}$ value is then compared to the critical value from the $\chi^{2}$ distribution table with degrees of freedom, df = (R - 1)(C - 1), and chosen confidence level. If the calculated $\chi^{2}$ value $>$ critical $\chi^{2}$ value, then we reject the null hypothesis. Calculating the $\chi^{2}$, we have sorted the feature variables in descending order of the chi-squared statistic values, $\chi^{2}$  and applied backward elimination method to find the key feature variables. First, we train the algorithms with all the feature variables and calculate the accuracy, then in each step we remove the feature variable with least $\chi^{2}$ value from the data-set and train the algorithms with one less feature variable and record the accuracy. Once we have recorded the accuracy scores for classifiers trained with $n$ ($n=(1,2,3...,36)$) best $\chi^{2}$ valued feature variables, we compare this scores between 36 classifiers and select the features of the best performing classifier as the key feature variables. We do this as the feature importance can play a role to build an effective machine learning based model \cite{sarker2020intrudtree}. After that we employ the machine learning classifiers mentioned in Section \ref{Introduction} to build the predictive binary classifier to achieve our goal.

\section{Evaluation and experimental results}
\label{Evaluation and experimental results}
\subsection{Evaluation}
In this study, we have calculated two important aspect of substance abuse: i. key features responsible for substance abuse, and ii. individual vulnerability. To identify the best performing features for the predictive classifier, the backward elimination method is applied. In our study, as test data is one-third of the whole data-set, accuracy of the prediction on the test data is a good metric to understand the classifier's performance on an unknown individual's data. For feature selection, we compare the accuracy scores of different classifiers trained with feature variables starting from all 36 feature variables coming all the way down to a single feature variable in 36 steps, excising the feature with least chi-squared statistic value in every step. Alongside accuracy, $ROC$(receiver operator characteristic) curve and the $AUC$(area under the curve) values for different classifiers are also calculated for comparing performance between predictive classifiers as in Figure \ref{fig:methodology_&_evaluation}, $AUC$ can be used as a criterion to measure the test's discriminating ability, i.e. how good is the test in a given clinical situation.

\subsection{Experimental results}
The first objective of this study is to identify the most influential risk factors of substance abuse. Features like $Belief\_in\_notion$,  $relation\_w\_parents$, $curiosity\_in\_drugs$, $friends\_no$ \& $life\_satis$, are the 5 features having the best chi-square statistic value. From our evaluation that follows, logistic regression classifier trained with 18 best $\chi^{2}$ features is identified as performing classifier. These 18 features are our key features of the initial 36, among this 18 feature variables, $belief\_in\_notion$, $life\_satis.$, $failed\_love$, and $curse\_for\_exp$ are based on depression and stress; $relation\_w\_parents$, $fam\_in\_drugs$, and $happiness\_in\_relationship$ are based on family and relations; $friends\_w\_drugs$, $friends\_no$, $party\_join$, $access\_to\_drugs$, and $hangout\_type$ are based on peer influence; $occupation\_v\_friends$, and $occupation\_succ$ are based on career failure and unemployment; $hate\_addict$, and $risk\_tendency$ are based on personality; $curiosity\_in\_drugs$ is based on curiosity; $religious\_mindset$ is based on religious affiliations. Insights from literature review and follow up with local psychiatrists as mentioned in the Introduction section support these findings.

As feature selection is performed using the backward elimination method, performance in each step is recorded to compare and analyze between the classifiers. Classifier having the best performance in each iteration is identified and then compared with other best performing models in consecutive iterations. Here, in Table \ref{tab:accuracy}, the accuracy scores of classifiers are displayed for 4 steps:

\begin{table}[!ht]
\caption{\% Accuracy of ML Classifiers trained with features selected by $\chi^{2}$ values}
\centering
 \begin{tabular}{||ccccc||} 
 \hline
  Classifier trained with &  17 features & 18 features & 19 features & all 36 features  \\ [0.5ex] 
 \hline\hline
 Random forest & 90.98 & 92.62 & 92.62 & 95.08 \\ 
 \hline
 KNearest Neighbors & 86.06 & 85.24 & 86.06 &88.52\\
 \hline
 Decision Tree & 85.24 & 86.88 & 88.52 & 85.24\\
 \hline
 Linear SVC & 95.08 & 95.08 & 95.08 & 95.90\\
 \hline
 Gaussian Naive Bayes & 91.80 & 91.80 & 90.16 & 92.62\\
 \hline
 Logistic Regression & 95.08 & 96.72 & 95.90 & 94.26\\ 
 \hline
 \end{tabular}
\label{tab:accuracy}
\end{table}
As displaying the results for all 36 steps is  difficult to accommodate and extravagant as well, here accuracy scores of the models trained with 17, 18, 19 and all 36 features respectively are displayed in Table \ref{tab:accuracy}. The best performance in each column is compared between all the columns resulting in the logistic regression classifier trained with 18 feature variables selected by chi-squared statistic value, $\chi^{2}$ having the highest accuracy score of 96.72\%. As train and test data has a ratio of 2:1 this score implies this classifier can accurately predict  162 individual data with 96.72\% accuracy.

Finally we compared the $ROC$ curves for each of the classifiers mentioned above in Table \ref{tab:accuracy}. 
\begin{figure}[h]
\centering
\begin{minipage}{.5\textwidth}
  \centering
  \includegraphics[width=\linewidth]{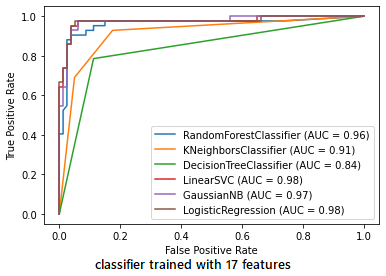}
  \label{fig:ROC_17features}
\end{minipage}%
\begin{minipage}{.5\textwidth}
  \centering
  \includegraphics[width=\linewidth]{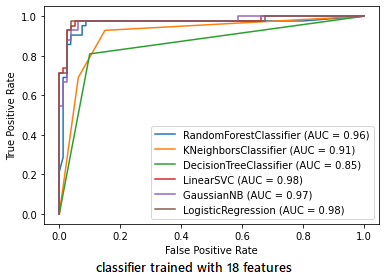}
  \label{fig:ROC_17features}
\end{minipage}
\begin{minipage}{.5\textwidth}
  \centering
  \includegraphics[width=\linewidth]{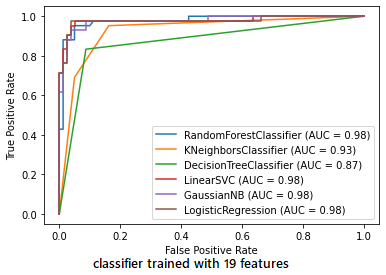}
  \label{fig:ROC_17features}
\end{minipage}%
\begin{minipage}{.5\textwidth}
  \centering
  \includegraphics[width=\linewidth]{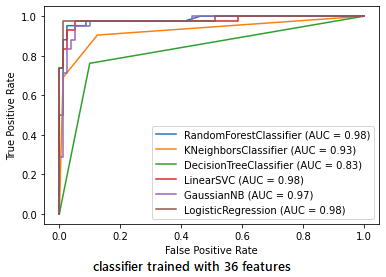}
  \label{fig:ROC_36features}
\end{minipage}
\caption{ROC curves for classifier trained with 17, 18, 19 and 36 best $\chi^{2}$ feature variables}
\label{fig:ROC_comparison}
\end{figure}
$AUC$  values for each classifier is displayed at the bottom right corner of the images. Calculating $AUC$ values together with accuracy scores gives us a convincing decision towards selection of our desired model. Analyzing the $AUC$ values in Figure \ref{fig:ROC_comparison}, we see Logistic Regression has the best combination with an accuracy score of 95.08\% with an $AUC$ value of 0.98 in case of classifiers trained with 17 features, for 18 features, Logistic Regression has the best combination with an accuracy score of 96.72\% with an $AUC$ value of 0.98, with 19 features, Logistic Regression has the best combination again with an accuracy of 95.90\% and an $AUC$ value of 0.98, finally when trained with all 36 features, LinearSVC has the best combination with an accuracy score of 95.90\% with an $AUC$ value of 0.98.

We have calculated accuracy score, $ROC$ curve and $AUC$ values and compared these values among all the classifiers to understand the effectiveness and finally compared the performances of all the classifiers. Performances of only four cases are shown in Table \ref{tab:accuracy} and Figure \ref{fig:ROC_comparison}, because it will be imprudent, redundant, and nearly impossible to accommodate all necessary aforementioned evaluation scores for each case with 36 feature variables meaning 36 possible cases. The performance of classifiers trained with less than 10 variables gradually declines with the exclusion of each feature variable in every step. Similarly, the classifiers trained with more than 20 variables has accuracy scores equal to or less than the ones trained with 20 feature variables which is evident in the performance of the classifier trained with all 36 feature variables in Table \ref{tab:accuracy}. Moreover, training classifiers with fewest possible feature variables not only reduces the time required for training but also eliminates the possibilities of over-fitting. Considering all these factors, Logistic regression classifier trained with 18 feature variables selected by highest chi-squared statistic value, $\chi^{2}$ is the most effective model to differentiate accurately between $healthy$ and $addicted$ classes with an accuracy score of 96.72\%, and $AUC$ value of 0.98 which can be considered as a skilled classifier regarding psychological diagnostics.

\section{Conclusion}
\label{Conclusion}
This study is performed among the young urban population in Dhaka city. Identifying key factors from literature review and discussing with professionals helped to prepare an effective questionnaire. We have collected data from the respondents of same age group and environment which have helped obtaining high accuracy. We have investigated the key factors and the vulnerability towards substance abuse of individuals who can subsequently be enlisted for counselling based on their predicted vulnerability score. Shifting geographical location can result in change in the key factors responsible for substance abuse, might affect the performance of the classifier. Subjects were from same background, performance can be slightly off while predicting subjects from diverse backgrounds, however collecting data from diverse backgrounds will make the classifier more inclusive. It leaves opportunity for future works: building automated counselling recommendation system, chat-bot and vulnerability index predictor.

\bibliographystyle{splncs04} 
\bibliography{predicting_vulnerability}

\begin{thebibliography}{10}
\providecommand{\url}[1]{\texttt{#1}}
\providecommand{\urlprefix}{URL }
\providecommand{\doi}[1]{https://doi.org/#1}

\bibitem{worldhealthorganization}
Mental health and substance abuse (Accessed: November 25,2020),
  \url{https://www.who.int/westernpacific/about/how-we-work/programmes/mental-health-and-substance-abuse}

\bibitem{aldarwish2017predicting}
Aldarwish, M.M., Ahmad, H.F.: Predicting depression levels using social media
  posts. In: 2017 IEEE 13th international Symposium on Autonomous decentralized
  system (ISADS). pp. 277--280. IEEE (2017)

\bibitem{bahr1998family}
Bahr, S.J., Maughan, S.L., Marcos, A.C., Li, B.: Family, religiosity, and the
  risk of adolescent drug use. Journal of Marriage and the Family pp. 979--992
  (1998)

\bibitem{barrett2006family}
Barrett, A.E., Turner, R.J.: Family structure and substance use problems in
  adolescence and early adulthood: examining explanations for the relationship.
  Addiction  \textbf{101}(1),  109--120 (2006)

\bibitem{brandon2007relapse}
Brandon, T.H., Vidrine, J.I., Litvin, E.B.: Relapse and relapse prevention.
  Annu. Rev. Clin. Psychol.  \textbf{3},  257--284 (2007)

\bibitem{conner2009meta}
Conner, K.R., Pinquart, M., Gamble, S.A.: Meta-analysis of depression and
  substance use among individuals with alcohol use disorders. Journal of
  substance abuse treatment  \textbf{37}(2),  127--137 (2009)

\bibitem{degenhardt2012extent}
Degenhardt, L., Hall, W.: Extent of illicit drug use and dependence, and their
  contribution to the global burden of disease. The Lancet  \textbf{379}(9810),
   55--70 (2012)

\bibitem{fisher1998predicting}
Fisher, L.A., Elias, J.W., Ritz, K.: Predicting relapse to substance abuse as a
  function of personality dimensions. Alcoholism: Clinical and Experimental
  Research  \textbf{22}(5),  1041--1047 (1998)

\bibitem{gowing2015global}
Gowing, L.R., Ali, R.L., Allsop, S., Marsden, J., Turf, E.E., West, R., Witton,
  J.: Global statistics on addictive behaviours: 2014 status report. Addiction
  \textbf{110}(6),  904--919 (2015)

\bibitem{hassanpour2019identifying}
Hassanpour, S., Tomita, N., DeLise, T., Crosier, B., Marsch, L.A.: Identifying
  substance use risk based on deep neural networks and instagram social media
  data. Neuropsychopharmacology  \textbf{44}(3),  487--494 (2019)

\bibitem{hawkins1992risk}
Hawkins, J.D., Catalano, R.F., Miller, J.Y.: Risk and protective factors for
  alcohol and other drug problems in adolescence and early adulthood:
  implications for substance abuse prevention. Psychological bulletin
  \textbf{112}(1), ~64 (1992)

\bibitem{heydarabadi2015prevalence}
Heydarabadi, A.B., Ramezankhani, A., Barekati, H., Vejdani, M., Shariatinejad,
  K., Panahi, R., Kashfi, S.H., Imanzad, M.: Prevalence of substance abuse
  among dormitory students of shahid beheshti university of medical sciences,
  tehran, iran. International journal of high risk behaviors \& addiction
  \textbf{4}(2) (2015)

\bibitem{kelley2002neuroscience}
Kelley, A.E., Berridge, K.C.: The neuroscience of natural rewards: relevance to
  addictive drugs. Journal of neuroscience  \textbf{22}(9),  3306--3311 (2002)

\bibitem{mchugh2013chi}
McHugh, M.L.: The chi-square test of independence. Biochemia medica: Biochemia
  medica  \textbf{23}(2),  143--149 (2013)

\bibitem{nagelhout2017economic}
Nagelhout, G.E., Hummel, K., de~Goeij, M.C., de~Vries, H., Kaner, E., Lemmens,
  P.: How economic recessions and unemployment affect illegal drug use: a
  systematic realist literature review. International Journal of Drug Policy
  \textbf{44},  69--83 (2017)

\bibitem{oetting1987peer}
Oetting, E.R., Beauvais, F.: Peer cluster theory, socialization
  characteristics, and adolescent drug use: A path analysis. Journal of
  counseling psychology  \textbf{34}(2), ~205 (1987)

\bibitem{pierce2005role}
Pierce, J.P., Distefan, J.M., Kaplan, R.M., Gilpin, E.A.: The role of curiosity
  in smoking initiation. Addictive behaviors  \textbf{30}(4),  685--696 (2005)

\bibitem{sani2010drug}
Sani, M.N.: Drug addiction among undergraduate students of private universities
  in bangladesh. Procedia-social and behavioral sciences  \textbf{5},  498--501
  (2010)

\bibitem{sarker2020intrudtree}
Sarker, I.H., Abushark, Y.B., Alsolami, F., Khan, A.I.: Intrudtree: A machine
  learning based cyber security intrusion detection model. Symmetry
  \textbf{12}(5), ~754 (2020)

\bibitem{sarker2020context}
Sarker, I.H., Alqahtani, H., Alsolami, F., Khan, A.I., Abushark, Y.B.,
  Siddiqui, M.K.: Context pre-modeling: an empirical analysis for
  classification based user-centric context-aware predictive modeling. Journal
  of Big Data  \textbf{7}(1),  1--23 (2020)

\bibitem{sarker2020mobile}
Sarker, I.H., Hoque, M.M., Uddin, M.K., Alsanoosy, T.: Mobile data science and
  intelligent apps: Concepts, ai-based modeling and research directions. Mobile
  Networks and Applications pp. 1--19 (2020)

\bibitem{sarker2020cybersecurity}
Sarker, I.H., Kayes, A., Badsha, S., Alqahtani, H., Watters, P., Ng, A.:
  Cybersecurity data science: an overview from machine learning perspective.
  Journal of Big Data  \textbf{7}(1),  1--29 (2020)

\bibitem{sarker2019effectiveness}
Sarker, I.H., Kayes, A., Watters, P.: Effectiveness analysis of machine
  learning classification models for predicting personalized context-aware
  smartphone usage. Journal of Big Data  \textbf{6}(1), ~57 (2019)

\bibitem{scholl2019drug}
Scholl, L., Seth, P., Kariisa, M., Wilson, N., Baldwin, G.: Drug and
  opioid-involved overdose deaths—united states, 2013--2017. Morbidity and
  Mortality Weekly Report  \textbf{67}(51-52), ~1419 (2019)

\bibitem{sinha2008chronic}
Sinha, R.: Chronic stress, drug use, and vulnerability to addiction. Annals of
  the new York Academy of Sciences  \textbf{1141}, ~105 (2008)

\end{thebibliography}
\end{document}